\pdfoutput=1

\documentclass[11pt]{article}

\usepackage[]{ACL2023}

\usepackage{times}
\usepackage{latexsym}

\usepackage[T1]{fontenc}

\usepackage[utf8]{inputenc}

\usepackage{microtype}

\usepackage{inconsolata}
\usepackage{hyperref}
\usepackage{url}

\usepackage{graphicx}
\usepackage{amsmath, bm}
\usepackage{multirow}
\usepackage{booktabs}
\usepackage{wrapfig}
\usepackage{sidecap}
\usepackage{arydshln}
\usepackage{amssymb}
\usepackage{color}
\usepackage{ulem}

\title{SpeechLM: Enhanced Speech Pre-Training with Unpaired Textual Data}

\author{Ziqiang Zhang$^{1,*}$, Sanyuan Chen$^{2,*}$, Long Zhou$^{3,}$\thanks{\ \ Equal contribution. The work was done at Microsoft Research Asia during the internship of the first two authors. Correspondence to:  Long Zhou (lozhou@microsoft.com)} \ , Yu Wu$^{3}$, Shuo Ren$^{3}$, Shujie Liu$^{3}$, \\ \textbf{Zhuoyuan Yao$^{3}$, Xun Gong$^{3}$, Lirong Dai$^{1}$, Jinyu Li$^{3}$, Furu Wei$^{3}$} \\
$^1$University of Science and Technology of China\\
$^2$Harbin Institute of Technology\\
$^3$Microsoft \\
}

\begin{document}

\maketitle

\begin{abstract}
How to boost speech pre-training with textual data is an unsolved problem due to the fact that speech and text are very different modalities with distinct characteristics. In this paper, we propose a cross-modal \textbf{Speech} and \textbf{L}anguage \textbf{M}odel (\textbf{SpeechLM}) to explicitly align speech and text pre-training with a pre-defined unified discrete representation. Specifically, we introduce two alternative discrete tokenizers to bridge the speech and text modalities, including phoneme-unit and hidden-unit tokenizers, which can be trained using a small amount of paired speech-text data. Based on the trained tokenizers, we convert the unlabeled speech and text data into tokens of phoneme units or hidden units. The pre-training objective is designed to unify the speech and the text into the same discrete semantic space with a unified Transformer network.
We evaluate SpeechLM on various spoken language processing tasks including speech recognition, speech translation, and universal representation evaluation framework SUPERB, demonstrating significant improvements on content-related tasks. 
Code and models are available at \url{https://aka.ms/SpeechLM}.
\end{abstract}

\section{Introduction}

Speech and text are two important carriers of human communication, and they can be converted into each other through speech recognition and synthesis systems.
In past years, the unimodal self-supervised representation learning has been well explored in natural language \citep{devlin2018bert,dong2019unified} and speech \citep{schneider2019wav2vec,hsu2021hubert}.
According to neuroscience, humans first pre-process speech and text with different cortices, and then extract the meaning with the same area, called the Wernicke-Geschwind area \citep{tremblay2016broca}.
Motivated by this,  it is a very promising direction to design two pre-nets and a unified representation space (similar to the Wernicke area) so that the speech model would benefit greatly from text modality.

In terms of joint speech-text modeling, most approaches employ a speech encoder and a text encoder to map the speech and text inputs to hidden states, based on which, a shared encoder is used to learn cross-modality content information \citep{bapna2021slam,bapna2022mslam,chen2022maestro}. 
To align the speech and text modalities, two alignment losses  (TLM and STM) in SLAM \citep{bapna2021slam} are introduced with supervised ASR data. 
Extending SLAM to the multilingual scenario, mSLAM \citep{bapna2022mslam} introduces CTC losses and uses SpanBERT \citep{joshi2020spanbert} to replace the BERT objective for pre-training on character-level text.
Based on the RNN-T framework, Maestro \citep{chen2022maestro} learns shared representations with modality matching, duration prediction, and sequence alignment.
Almost all previous work follows the same structure with a speech/text encoder and a shared encoder, however, the interface between the speech encoder and the text encoder is not well studied, which probably leads to the outputs of the two encoders in different spaces, and suffers from transfer interference and capacity dilution for the shared encoder \citep{bapna2021slam}.

In this paper, we aim at unifying speech and text modalities via a well-defined interface, with which the model can benefit from additional textual data.
We argue that such an interface should provide a shared semantic space for both speech and text, and preferably have strong interpretability and learnability.
To this end, we explore two alternative representation spaces satisfying the above characteristics of the interface, which are based on phoneme units and hidden units.
Specifically, we introduce two discrete tokenizers named \textbf{phoneme-unit tokenizer} and \textbf{hidden-unit tokenizer}. All tokenizer models are obtained with unsupervised data or a small amount of ASR data and are used offline before pre-training.
With them, we convert the speech and text to a shared intermediate modality (phoneme/hidden units), and decouple the joint speech-text modeling into two sub-modules, i.e., the learning of mapping between speech/text and the discrete units.
Specifically, we propose two pre-training tasks. One is \textbf{Unit-based Masked Language Modeling (UMLM)} trying to predict the unit tokens from the masked speech.
The other one is \textbf{Unit-based Connectionist Temporal Classification (UCTC)} task, aiming at reconstructing the whole text sequences from the masked unit sequences.
To better align the representations of speech and text, we also adopt a \textbf{Random Swapping Mechanism} for the UMLM task, swapping the intermediate representations of the speech and the corresponding discrete units before feeding them into the shared subsequent network.

The contributions of this paper are summarized as follows.
\begin{itemize}
\item We propose two alternative tokenizers which can convert unlabeled speech and text into the shared discrete space and relieve the influence of modality difference.
\item The proposed SpeechLM (\textbf{Speech} and \textbf{L}anguage \textbf{M}odel), equipped with two simple and clear learning objectives and the random swapping mechanism, can unify and simplify the cross-modal speech-text pre-training.
\item Experiments demonstrate that SpeechLM enhanced by textual data significantly outperforms its speech-only counterparts on various spoken language tasks, e.g., ASR, speech translation (ST), and universal representation evaluation framework SUPERB \citep{yang2021superb}.
\end{itemize}


\section{Related Work}

\paragraph{Predictive Representation Learning for Speech}
Unlike natural language processing (NLP), speech signals are continuous, making it not straightforward to find the predictive labels for pre-training. To tackle this issue, a tokenizer, also referred to as a quantizer, is required to map continuous speech features into discrete tokens \citep{vq_wav2vec,hsu2021hubert,chung2021w2v}.
HuBERT \citep{hsu2021hubert} is the pioneer in the exploration of predictive speech representation learning (SSL), which utilizes a k-means model on the middle layer of the Transformer as the tokenizer to convert speech into discrete tokens.
\citet{chung2021w2v} tries to combine a contrastive loss and a masked prediction loss in a self-supervised speech representation learning framework.
In addition to the unsupervised tokenizers, \citet{wang2022supervision} proposes a supervision-guided tokenizer, which is an acoustic model trained on limited labeled data, and can generate frame-level aligned phonemes as the predictive targets for SSL.
In contrast, our goal is to take advantage of textual data to improve speech representation learning.

\begin{figure*}[t]
	\centering
	\includegraphics[width=0.95\textwidth]{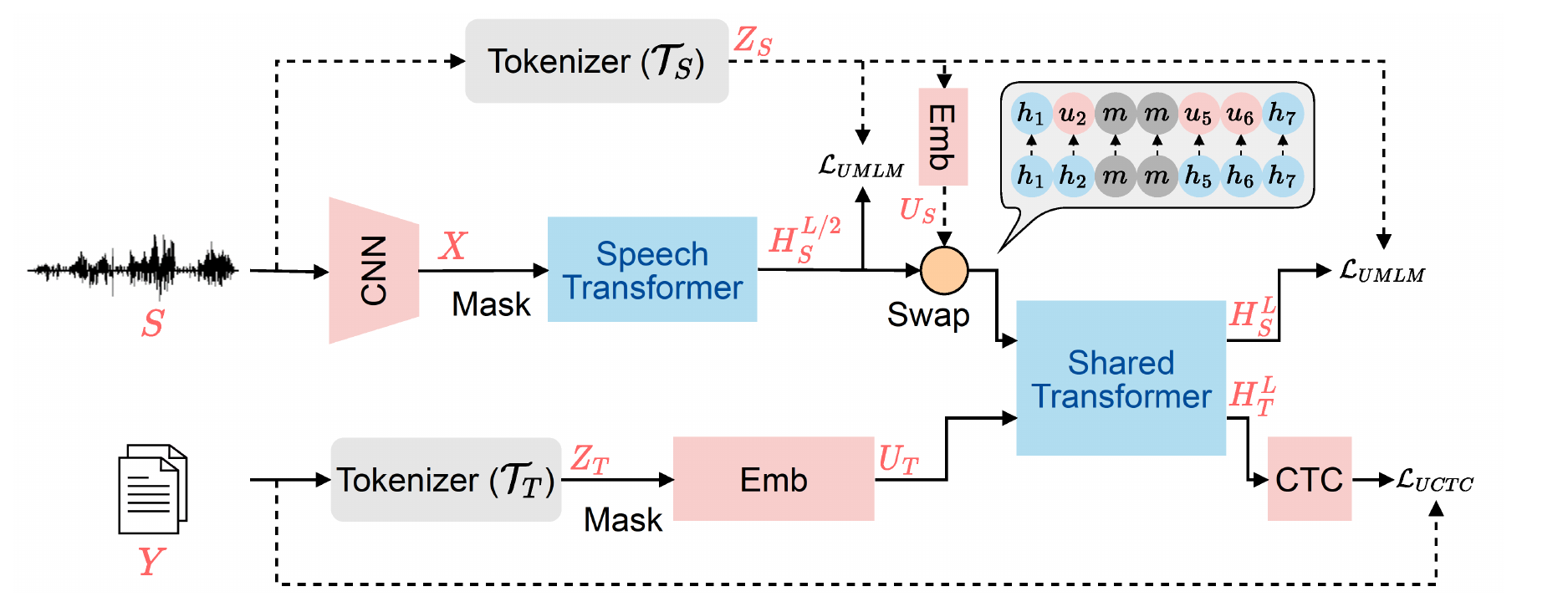}
	\caption{SpeechLM pre-training framework, which consists of a Speech Transformer and a Shared Transformer, and equips with discrete tokenizers and the random swapping mechanism.}\label{architecture}
\end{figure*}%

\paragraph{Joint Speech-Text Modeling}
With the rapid development of unimodal pre-training in speech and natural language processing \citep{devlin2018bert,hsu2021hubert}, joint speech-text pre-training obtains more and more attention from research and industrial communities \citep{kim2021st,qian2021speech,bapna2021slam,ao2021speecht5,tang2022unified,zhang2022speechut}.
Most previous studies \citep{kim2021st,qian2021speech,ao2021speecht5} let speech and text share some parameters of a neural network in pre-training, however, the speech and the text are not guaranteed to lie in the same space, suffering from transfer interference and capacity dilution \citep{bapna2021slam}.
To alleviate this issue, SLAM \citep{bapna2021slam} and mSLAM \citep{bapna2022mslam}, which are most related work to our SpeechLM, leverage extra supervised speech-to-text tasks to enhance the speech-text alignment.
However, these approaches still leave unpaired speech and text data modeled separately by using different pre-training targets, which might lead the model to use individual capacities to handle each modality, and can not guarantee speech and text lie in the same space.

Our work is also related to MAESTRO \citep{chen2022maestro}, which learns shared representations from speech and text modalities with a modality matching algorithm in RNN-T framework, but the modality matching could be only performed on a few paired speech-text data, limiting the effectiveness of alignment learning. 
Unlike SLAM and MAESTRO, we utilize trained tokenizers to convert all unpaired speech and text into the same discrete space and eliminate the influence of modal difference, so that the two modalities can interact naturally via the shared interface during the pre-training.
SpeechUT \citep{zhang2022speechut} also leverages hidden units as the bridge of speech and text modalities, while it only works in an encoder-decoder architecture where the encoder mainly models the speech and the decoder mainly models the text.
Instead, this work focuses on the encoder architecture which is more suitable for general speech representations, and explores different tokenizers and pre-training tasks.

\section{Methods}
Given unpaired speech and text data, SpeechLM is pre-trained to learn a unified representation of speech and text modalities with the help of offline discrete tokenizers.
In this section, we will present the overall framework of SpeechLM, as well as the pre-training procedures and the tokenizers.

\subsection{Phoneme/Hidden Unit as the Bridge}
Speech and language are two different modalities with different characteristics.
We explore bridging speech and text pre-training with an explicitly defined discrete representation, where speech and text could be tokenized into a shared discrete space easily.
Leveraging phoneme/hidden units as the bridge between speech and text has the following advantages:
First, it is easier to separately align speech and text into a shared intermediate representation than to align them directly.
Second, we can make full use of additional unpaired data to improve the alignment;
Thirdly, we can leverage more fine-grained alignment information, i.e., at the frame level, to facilitate joint modeling.

To achieve this goal, we implement two tokenizers for both speech and text, a phoneme-unit tokenizer and a hidden-unit tokenizer, which will be described in detail in Section \ref{Tokenizers}.
The former aims to convert speech and text into the phoneme space, while the latter converts them into an acoustic clustering space.
Given a speech sample $\bm{S}$ or a text sample $\bm{Y}$, a tokenizer ($\mathcal{T}_S$ for speech, $\mathcal{T}_T$ for text) yields a sequence of discrete units $\bm{Z}$,
\begin{equation}
\begin{aligned}
    \bm{Z}_S & \triangleq (z_{S_1}, \ldots , z_{S_M}) = \mathcal{T}_S(S), \\
    \bm{Z}_T & \triangleq (z_{T_1}, \ldots , z_{T_N}) = \mathcal{T}_T(Y)
    \label{eqn:tokenize}
\end{aligned}
\end{equation}
where $M$ and $N$ are the lengths of the unit sequences from speech and text, respectively.

\subsection{Model Architecture}
SpeechLM consists of a Speech Transformer and a Shared Transformer, which are enhanced with the random swapping mechanism, as illustrated in Figure \ref{architecture}.
Next, we will introduce the main modules with the input of unpaired speech $\bm{S}$ and text $\bm{Y}$.

\paragraph{Speech Transformer}
Following HuBERT \citep{hsu2021hubert}, we use a standard Transformer \citep{vaswani2017attention} as the backbone of the Speech Transformer, equipped with relative position embedding \citep{shaw-etal-2018-self}.
A speech waveform $\bm{S}$ is first processed into a sequence of speech features $\bm{X} \triangleq (x_1, x_2, \dots, x_{M})$ by a stack of 1-D convolutional layers.
We follow HuBERT to mask the speech feature $\bm{X}$ with the mask probability of 8\% and the mask length of 10.
Then the masked features, $\bm{\hat{X}}$, are fed into the Speech Transformer for higher-level representations,
\begin{equation}
    \bm{H}_S^{l}={\rm{Transformer}}(\bm{H}_S^{l-1})
    \label{eqn:speech_h}
\end{equation}
where $l$ means the layer and $\bm{H}_S^{0} \triangleq \bm{\hat{X}}$ indicating the input.
Let $L$ be the total number of layers of all Transformer modules and the Speech Transformer accounts for half, consequently, the output is $\bm{H}_S^{L/2} \triangleq (h_{S_1}^{L/2},..., h_{S_M}^{L/2})$.

\paragraph{Shared Transformer}
The Shared Transformer has the same architecture with the Speech Transformer and handles two types of input with respect to speech and text.
The first input is the previous output of the Speech Transformer, $\bm{H}_S^{L/2}$, and it is processed by the Shared Transformer into $\bm{H}_S^{L}$.
The second input is the unit embedding sequence $\bm{U}_T \triangleq (u_{T_1}, \dots, u_{T_N})$ that is derived from the text tokenized units $\bm{Z}_T$ by the unit embedding layer,
\begin{equation}
    \bm{U}_T = \mathrm{Emb}(\bm{Z}_T)
    \label{eqn:emb_t}
\end{equation}
It is then processed by the Shared Transformer into $\bm{H}_T^{L}$, where $\bm{H}_T^{L/2} \triangleq \bm{U}_T$ indicates the input.
Consequently, $\bm{H}_S^{L}$ and $\bm{H}_T^{L}$ are used as the encoded representations for speech and text.
For textual representations, we further employ a CTC layer \citep{Graves10.1145CTC} that converts $\bm{H}_T^{L}$ to character-level representations.

\paragraph{Random Swapping Mechanism}
To better align the speech and textual representations into shared latent space at the early layer of the Shared Transformer, we introduce a random swapping mechanism.
As each speech sequence can be tokenized into discrete units, we can randomly select some time positions (denoted as $i\in \mathcal{R}$) from a speech sequence and replace each $h_{S_i}^{L/2}$ with the corresponding unit embedding ${u_{S_i}}$, where ${u_{S_i}} = \mathrm{Emb}(z_{S_i})$ is derived from speech units $z_{S_i}$ by the unit embedding layer.
To avoid information leakage, the swapping positions $\mathcal{R}$ are only selected within unmasked regions of speech sequence.
In this way, we can shuffle two modalities into one sequence and the model can treat them equally.

\subsection{Pre-Training Tasks}
SpeechLM is jointly optimized by a unit-based masked language modeling task with unlabeled speech data and a unit-based connectionist temporal classification task with unlabeled text data.

\paragraph{Unit-based Masked Language Modeling (UMLM)}
The unit-based masked language modeling task is designed for speech pre-training, like HuBERT \citep{hsu2021hubert} and ILS-SSL \citep{wang2021self}.
Given $l$-layer speech representations $\bm{H}_S^{l} \triangleq (h^{l}_{S_1}, \ldots, h^{l}_{S_M})$, UMLM tries to predict the corresponding tokenized units $\bm{Z}_S \triangleq (z_{S_1}, \ldots , z_{S_M})$ at the masked positions.
The probability of the predicted unit at position $i$ is calculated with
\begin{equation}
\label{sim_compute}
    p(z|{h^{l}_S}_i) = \frac{\exp(\text{cos}(\bm{W}h^{l}_{S_i}, \bm{e}(z))/\tau)}{\sum_{z^{'}\in \mathcal{Z}} \exp(\text{cos}(\bm{W}h^{l}_{S_i}, \bm{e}(z^{'}))/\tau)}
\end{equation}
where $\bm{W}$ is a projection matrix, $\bm{e}(.)$ is an embedding matrix, $\tau$ = 0.1 is the temperature coefficient, and $\mathcal{Z}$ is the set of phoneme/hidden-unit categories.
Similar to ILS-SSL, the UMLM loss is computed on both the outputs of Speech Transformer ($\bm{H}_S^{L/2}$) and Shared Transformer ($\bm{H}_S^{L}$), with the loss formulated as,
\begin{equation}
\begin{aligned}
    & \mathcal{L}_{{\rm{UMLM}}} = \\
    & - \sum_{i\in \mathcal{M}}\left(\text{log}{~p(z_{S_i}|h_{S_i}^{L/2})} + \text{log}{~p(z_{S_i}|h^{L}_{S_i})}\right)
    \label{eq:UMLM}
\end{aligned}
\end{equation}
where $z_{S_i}$ is the corresponding speech unit at position $i$ and $\mathcal{M}$ is the set of masked positions.

\paragraph{Unit-based Connectionist Temporal Classification (UCTC)}
Connectionist temporal classification (CTC) \citep{Graves10.1145CTC} is first proposed to address the sequence label problem where the output is shorter than the unsegmented input sequences.
Here, we take the phoneme-unit or hidden-unit sequences $\bm{Z}_T$ tokenized and upsampled from the unlabeled text as the input, and aim at recognizing the original text through the Shared Transformer and CTC layer.
The input sequence is masked in the same way as the input of the speech signal.
Given a text label sequence $\bm{Y}$, the unit-based CTC loss is calculated as,
\begin{equation} \label{ctc}
    \mathcal{L}_{{\rm{UCTC}}} = - \text{log}\  p_{CTC}(\bm{Y} | \bm{H}_T^{L})
\end{equation}
where $p_{CTC}(\cdot)$ is modeled by the CTC layer, whose goal is to transform the encoded unit representation $\bm{H}_T^{L}$ into the target characters $\bm{Y}$.

By taking advantage of  unlabeled speech and text data, SpeechLM performs multi-task pre-training with UMUM and UCTC tasks,
\begin{equation} \label{total_loss}
    \mathcal{L} = \mathcal{L}_{{\rm{UMUM}}} + \lambda \mathcal{L}_{{\rm{UCTC}}}
\end{equation}
where $\lambda$ is used to control the weight of two losses.
Through joint optimization and the random swapping mechanism, SpeechLM is expected to align speech and text into a unified representation.

\subsection{Unified Tokenizers} \label{Tokenizers}
\begin{figure}[ht]
    \centering
    \includegraphics[width=1.0\columnwidth]{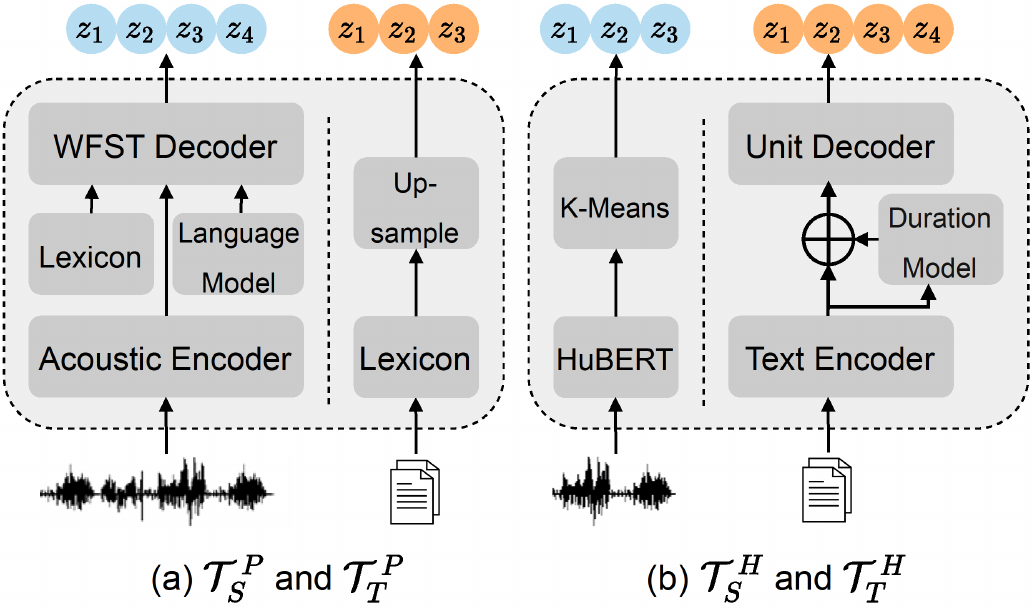}
    \caption{Two alternative tokenizers for speech and text, (a) Phone-unit tokenizer; (b) Hidden-unit tokenizer.}
    \label{fig:toks}
\end{figure}

Figure \ref{fig:toks} shows the overview of the proposed phoneme-unit tokenizer and hidden-unit tokenizer. 
Besides, the tokenizers are offline models, which are used to pre-process the unlabeled speech and text data before the pre-training.

\paragraph{Phoneme-Unit Tokenizer}
Inspired by PBERT \citep{wang2022supervision}, which leverages phoneme labels as the pre-training targets, we introduce the phoneme-unit tokenizer ($\mathcal{T}^P$) to discretize speech signals ($\mathcal{T}^P_S$) as well as text sequences ($\mathcal{T}^P_T$).
For speech data, the tokenizer is composed of an acoustic model, whose goal is to convert acoustic features into phoneme units through a weight finite-sate transducer (WFST) based decoder \citep{mohri2002weighted}.
We implement it using the open-source Kaldi toolkit\footnote{https://github.com/kaldi-asr/kaldi} with a small amount of paired ASR data and language model (LM) data, with details described in Appendix \ref{Appen:tok_detail} due to the space limitation.
For text data, we can directly convert words into phonemes by looking up the provided lexicons.
We further upsample the phoneme sequences of text by randomly repeating each phoneme many times to make sure they have similar lengths to the phoneme sequences of speech.

\paragraph{Hidden-Unit Tokenizer}
We follow HuBERT to tokenize speech into hidden units with a k-means cluster model, called $\mathcal{T}^H_S$, where the clustering feature is the intermediate hidden states of the 2nd round HuBERT model.
Inspired by \citet{zhang2022speechut}, to tokenize text data into the same hidden-unit space, we propose a non-autoregressive text to hidden-unit model ($\mathcal{T}^H_T$), which is based on FastSpeech \citep{ren2019fastspeech}. The model consists of a text encoder, a duration model, and a unit decoder, as shown in Figure \ref{fig:toks} (b).
$\mathcal{T}^H_T$ is trained with a small amount of text-to-unit pairs from ASR data, where the text side is the phoneme transcriptions with phoneme's durations, and the units are tokenized from the corresponding speech by $\mathcal{T}^H_S$.
At inference time, $\mathcal{T}^H_T$ only consumes non-aligned phoneme sequences converted from raw text since the duration is automatically estimated.

\section{Experiment}
\label{experiment}
SpeechLM is evaluated on various spoken language tasks, including automatic speech recognition (ASR), speech translation (ST), and the universal representation evaluation benchmark SUPERB \citep{yang2021superb}.
According to the tokenizers, the model can be divided into \textbf{SpeechLM-H} and \textbf{SpeechLM-P} using hidden-unit and phone-unit as discrete tokens, respectively.

\subsection{Data}
We use unlabeled speech data from LibriSpeech \citep{panayotov2015librispeech} and LibriLight \citep{kahn2020librilight} to pre-train \textbf{Base} and \textbf{Large} models respectively.
LibriSpeech contains 960 hours of labeled speech where the labels are not used in pre-training.
LibriLight has about 60,000 hours of unlabeled speech in the same domain as LibriSpeech.
The unpaired text data are from LibriSpeech LM corpus\footnote{http://www.openslr.org/11/}, containing about 40M English sentences.
The paired data for optimizing the tokenizers are the full LibriSpeech data in the \textbf{Large} setting and the 100-hour subset (\texttt{train-clean-100}) in the \textbf{Base} setting.
For downstream tasks, we use LibriSpeech for ASR evaluation, and four translation directions of CoVoST-2 \citep{wang2020covost} for ST evaluation.
For all tasks of SUPERB evaluation, the data details can be found in \citet{yang2021superb}.

\begin{table*}[t!]
\begin{center}
\resizebox{\linewidth}{!}{
\renewcommand\arraystretch{1.0}
\begin{tabular}{lc|ccc|cc|ccc}
\hline
\multirow{2}{*}{Model}                  & \multirow{2}{*}{Size}  & \multicolumn{3}{c|}{Pre-training Data}        & \multicolumn{2}{c|}{WER ($\downarrow$) w/o LM} & \multicolumn{3}{c}{WER ($\downarrow$) w/ LM}   \\
                                        &                        & Speech           & Paired         & Text      & test-clean          & test-other         & LM        & test-clean& test-other \\
\hline
\multicolumn{10}{l}{\textit{100h fine-tuned}} \\
\hline
Wav2vec 2.0 \citep{wav2vec2}            & Base (0.1B)            & 960h             & -               & -           & 6.1                & 13.3             & 4-gram    & 3.4       & 8.0 \\
HuBERT \citep{hsu2021hubert}            & Base (0.1B)            & 960h             & -               & -           & 6.3                & 13.2             & 4-gram    & 3.4       & 8.1 \\
WavLM \citep{wavlm}                     & Base (0.1B)            & 960h             & -               & -           & 5.7                & 12.0             & 4-gram    & 3.4       & 7.7 \\
PBERT \citep{wang2021self}              & Base (0.1B)            & 960h             & 100h            & -           & 4.2                & 9.5              & 4-gram    & 3.1       & 7.2 \\
ILS-SSL \citep{wang2021self}            & Base (0.1B)            & 960h             & -               & -           & 4.7                & 10.1             & 4-gram    & 3.0       & 6.9 \\
data2vec \citep{baevski2022data2vec}    & Base (0.1B)            & 960h             & -               & -           & 4.2$^\ast$         & 9.7$^\ast$       & 4-gram    & 2.8       & 6.8 \\
\textbf{SpeechLM-H}                & Base (0.1B)            & 960h    & 100h    & 400K                          & 3.8  & 8.3 & 4-gram   &  \textbf{2.7}          &  \textbf{6.0}  \\
\textbf{SpeechLM-P}                     & Base (0.1B)            & 960h    & 100h    & 40M                      & \textbf{3.4}       & \textbf{8.1}  & 4-gram   &  \textbf{2.7}  &  6.2  \\
\hline
\multicolumn{10}{l}{\textit{960h fine-tuned}} \\
\hline
Wav2vec 2.0 \citep{wav2vec2}            & Large (0.3B)          & 60kh        & -           & -                     & 2.2                & 4.5               & Transf.      & 1.8            & 3.3            \\
HuBERT \citep{hsu2021hubert}            & Large (0.3B)          & 60kh        & -           & -                     & 2.1$^\ast$         & 4.2$^\ast$        & Transf.      & 1.9            & 3.3            \\
WavLM \citep{wavlm}                     & Large (0.3B)          & 94kh        & -           & -                     & -                  & -                 & Transf.      & 1.8            & 3.2            \\
ILS-SSL \citep{wang2021self}            & Large (0.3B)          & 60kh        & -           & -                     & 1.9                & 3.8               & Transf.      & 1.8            & 3.2            \\
\textbf{SpeechLM-P}                     & Large (0.3B)          & 60kh        & 960h        & 40M                   & \textbf{1.9}       & \textbf{3.6}      & Transf.      & \textbf{1.8}   & \textbf{3.2}   \\
\hline
\end{tabular}
}
\caption{ASR performance (WER) of different pre-trained models on the LibriSpeech benchmark. \textbf{Speech/Text} indicates the unpaired speech and text data, \textbf{Paired} indicates the paired ASR data for building tokenizers instead of directly used in pre-training. $^\ast$ indicates our reproduction results.
}
\label{Tab: wer_librispeech} 
\end{center}
\end{table*}

\subsection{Pre-Training Setup}
The network architecture of SpeechLM follows that of HuBERT \citep{hsu2021hubert} for a fair comparison.
Specifically, the \textbf{Base} model consists of $L$=12 Transformer layers where both the Speech Transformer and the Shared Transformer have 6 layers.
The \textbf{Large} model doubles the number of Transformer layers.
The convolutional layers downsample the input waveform to a frame rate of 20ms.
The CTC layer consists of a single 1-D convolutional layer followed by a linear layer, which outputs the probabilities of text characters.
All models are pre-trained on 32 GPUs for 400K steps.
To align with HuBERT, the update frequency is set to 4 for \textbf{Large} models to simulate 128 GPUs.
The batch size for the \textbf{Base} model is 4375 tokens after down(up)-sampling for both speech and text input, and for the \textbf{Large} model it is set to 2800.
The text loss ($\mathcal{L}_{UCTC}$) is weighted by 0.1\footnote{The effect of different weights ($\lambda$) is reported in Appendix \ref{Appen:pretraining_ratio}.}.
More details about the model configuration and training details can be found in Appendix \ref{exp_detail}.

\subsection{Evaluation on Speech Recognition}
We first verify the pre-trained SpeechLM on ASR tasks, where the Speech Transformer, the Shared Transformer, and the CTC head are fine-tuned with a speech-to-text CTC loss.
Base models are fine-tuned on the \texttt{train-clean-100} subset and Large models are fine-tuned on the full 960h LibriSpeech.
We measure the quality of ASR by the word error rate (WER) evaluated on the standard \texttt{test-clean/other} sets.
Table \ref{Tab: wer_librispeech} shows that in the Base setting, by taking advantage of textual data, SpeechLM significantly outperforms previous models, such as wav2vec 2.0 \citep{wav2vec2}, HuBERT \citep{hsu2021hubert}, and data2vec \citep{baevski2022data2vec}.
Particularly, the proposed SpeechLM obtains 26\% and 12\% relative WER reductions over HuBERT and data2vec on \texttt{test-other} set, respectively.
We notice that using 400K instead of the full 40M text data is better for SpeechLM-H models, as discussed later in \ref{ssec:analysis}.
Furthermore, our SpeechLM Large model achieves competitive or even better performance than previous work\footnote{SLAM and MAESTRO use 2$\times$ model size, larger amount of paired data, or different inference framework (e.g., RNN-T in MAESTRO), whose results (see Appendix \ref{Appen:XLarge-asr}) are not comparable with the setting in Table \ref{Tab: wer_librispeech}.}.

\subsection{Evaluation on Speech Translation}
We then evaluate SpeechLM on speech-to-text translation tasks.
Following \citet{Changhan2021Large}, we use four language directions from English to German (de), Catalan (ca), Arabic (ar), and Turkish (tr) in CoVoST-2 \citep{wang2020covost}.
When fine-tuning, the pre-trained model serves as the encoder, followed by a randomly initialized decoder consisting of 6 Transformer layers with a model dimension of 768.
We use character vocabulary for target languages in all translation tasks, and report the case-sensitive detokenized BLEU \citep{papineni2002bleu} on the test set.
The results are shown in Table \ref{Tab: ST_result}, including the baselines that are fine-tuned from other pre-trained models.
The numbers in brackets represent the standard deviation of three fine-tuning results.
Table \ref{Tab: ST_result} shows that by boosting the quality of speech representation learning with textual data, SpeechLM-H and SpeechLM-P achieve comparable results in the Base setting, with 2.4 BLEU improvement over HuBERT Base.
Surprisingly, the SpeechLM-P Large model substantially outperforms previous work with a smaller encoder, such as SLAM X-Large \citep{bapna2021slam}.

\begin{table*}[!htp]
\begin{center}
\small
\renewcommand\arraystretch{1.05}
\begin{tabular}{l|l|cccc|c}
\hline 
Pre-trained Model                                    & Encoder Size   & en-de & en-ca & en-ar & en-tr & avg  \\
\hline 
Pre-ASR \citep{wang2020covost}                       & -                & 16.3  & 21.8  & 12.1  & 10.0  & 15.1 \\
HuBERT \citep{hsu2021hubert} $^\ast$                 & Base (0.1B)      & 21.6  & 28.4  & 15.9  & 14.4  & 20.1 \\
\textbf{SpeechLM-H}                                  & Base (0.1B)      & 23.9 (0.2)  & 30.8 (0.1)	 & 18.0  (0.2)  & 16.1 (0.1)	 & 22.2 \\
\textbf{SpeechLM-P}                                  & Base (0.1B)      & \textbf{24.3} (0.1)  & \textbf{31.1} (0.1)  & \textbf{18.3} (0.1)  & \textbf{16.2} (0.1)  & \textbf{22.5} \\
\hline 
wav2vec 2.0 \citep{Changhan2021Large}                & Large (0.3B)     & 23.8  & 32.4  & 17.4  & 15.4  & 22.3 \\
SLAM \citep{bapna2021slam}                           & X-Large (0.6B)   & 27.2  & 33.3  & 18.5  & 16.8  & 24.0 \\
SLAM$\rightarrow$w2v-bert \citep{bapna2021slam}      & X-Large (0.6B)   & 27.1  & 34.2  & 21.2  & 17.5  & 25.0 \\ 
\textbf{SpeechLM-P}                                  & Large (0.3B)     & \textbf{27.3} (0.3)  & \textbf{35.9} (0.1) & \textbf{21.7} (0.4)  & \textbf{19.7} (0.3)  & \textbf{26.2} \\
\hline 
\end{tabular}
\caption{\label{Tab: ST_result} BLEU scores on four translation tasks of CoVoST-2, comparing SpeechLM with previous self-supervised models. $^\ast$ indicates our reproduction results.}
\end{center}
\end{table*}

\begin{table*}[ht]
\centering
\resizebox{1.0\textwidth}{!}{
\renewcommand\arraystretch{1.1}
\begin{tabular}{l|c|c||r|r|r||r|r|r|r|r||r|r|rr||r}
\hline
\multirow{3}{*}{Method} & \multirow{3}{*}{\#Params} & \multirow{3}{*}{Corpus} & \multicolumn{3}{c||}{Speaker} & \multicolumn{5}{c||}{Content} & \multicolumn{4}{c||}{Semantics}  & ParaL  \\ \cline{4-16}

& & & SID & ASV & SD & PR & ASR & OOD-ASR & KS & QbE & ST & IC& \multicolumn{2}{c||}{SF} & ER   \\ \cline{4-16}

& & & Acc $\uparrow$ & EER $\downarrow$ & DER $\downarrow$ & PER $\downarrow$ & WER $\downarrow$ & WER $\downarrow$ & Acc $\uparrow$ & MTWV $\uparrow$  & BLEU $\uparrow$ & Acc $\uparrow$ & F1 $\uparrow$ & CER $\downarrow$ & Acc $\uparrow$  \\ \hline 

FBANK & 0 & - & 8.5E-4 & 9.56 & 10.05 & 82.01 & 23.18 & 63.58 & 8.63 & 0.0058 &  2.32 & 9.10 & 69.64 & 52.94 & 35.39    \\ \hline

PASE+~\citep{pase+} & 7.83M & LS 50 hr & 37.99 & 11.61 & 8.68 & 58.87 & 25.11 & 61.56 & 82.54 & 0.0072 &  3.16 & 29.82 & 62.14 & 60.17 & 57.86   \\ \hline

APC~\citep{apc1} & 4.11M & LS 360 hr & 60.42 & 8.56 & 10.53 & 41.98 & 21.28 & 63.12  & 91.01 & 0.0310 & 5.95 & 74.69 & 70.46 & 50.89 & 59.33  \\

VQ-APC~\citep{vq_apc} & 4.63M & LS 360 hr & 60.15 & 8.72 & 10.45 & 41.08 & 21.20 & 63.56 & 91.11 & 0.0251 & 4.23 & 74.48 & 68.53 & 52.91 & 59.66  \\

NPC~\citep{npc} & 19.38M & LS 360 hr & 55.92 & 9.40 & 9.34 & 43.81 & 20.20 & 61.66  & 88.96 & 0.0246 &  4.32 & 69.44 & 72.79 & 48.44 & 59.08    \\

Mockingjay~\citep{mockingjay} & 85.12M & LS 360 hr & 32.29 & 11.66 & 10.54 & 70.19 & 22.82  & 65.27 & 83.67 & 6.6E-04 & 4.45 & 34.33 & 61.59 & 58.89 & 50.28   \\

TERA~\citep{tera} & 21.33M & LS 960 hr & 57.57 & 15.89 & 9.96 & 49.17 & 18.17  & 58.49  & 89.48 & 0.0013 & 5.66  & 58.42 & 67.50 & 54.17 & 56.27 	 \\

DeCoAR 2.0~\citep{decoar2} & 89.84M & LS 960 hr & 74.42 & 7.16 & 6.59 & 14.93 & 13.02 & 53.62 & 94.48 & 0.0406 & 9.94 & 90.80 & 83.28 & 34.73 & 62.47    \\

\hline

modified CPC~\citep{modified_cpc} & 1.84M & LL 60k hr & 39.63 & 12.86 & 10.38 & 42.54 & 20.18 & 62.54  & 91.88 & 0.0326 &  4.82 & 64.09 & 71.19 & 49.91 & 60.96 \\

wav2vec~\citep{schneider2019wav2vec} & 32.54M & LS 960 hr & 56.56 & 7.99 & 9.9 & 31.58 & 15.86 & 55.86  & 95.59 & 0.0485 &  6.61 & 84.92 & 76.37 & 43.71 & 59.79    \\

vq-wav2vec~\citep{vq_wav2vec} & 34.15M & LS 960 hr & 38.80 & 10.38 & 9.93 & 33.48 & 17.71 & 60.66 & 93.38 & 0.0410 & 5.66 & 85.68 & 77.68 & 41.54 & 58.24  \\

Wav2vec 2.0 Base~\citep{wav2vec2} & 95.04M & LS 960 hr & 75.18  & 6.02 & 6.08 & 5.74 & 6.43 & 46.95 & 96.23 & 0.0233 & 14.81 & 92.35 & 88.30 & 24.77 & 63.43 	 \\

HuBERT Base~\citep{hsu2021hubert} & 94.68M & LS 960 hr & 81.42 & 5.11 & 5.88  & 5.41 & 6.42 &46.69 & 96.30 & 0.0736 & 15.53 & 98.34 & 88.53 & 25.20 & 64.92 	 \\

WavLM Base \citep{wavlm} &   94.70M & LS 960 hr &84.51&	4.69&	4.55 &	4.84&	6.21	&	42.81 	& 96.79 & 0.0870 & 20.74 & 98.63 &	89.38 &	22.86 & 65.94 \\


\hline


\textbf{SpeechLM-H} Base &  94.70M & LS 960 hr & 76.90 & 5.79 &	6.10 &	3.70 &	\textbf{4.85}   	&	45.82 &	95.91  & 0.0485 	& \textbf{21.90}  &   98.52	& 88.80 & 24.20	  &	63.77    \\

\textbf{SpeechLM-P} Base &  94.70M & LS 960 hr & 75.24 & 5.97   &	7.34 &	\textbf{3.10} &	4.98   	&	49.04 &	94.09  & 0.0410 	& 19.20  &   97.68	& 87.67  & 25.90  &	61.84    \\





\hline
\end{tabular}}
\caption{\label{table:superb} Universal speech representation evaluation on the SUPERB benchmark with 12 tasks. 
}
\end{table*}

\subsection{Universal Representation Evaluation}
We further evaluate our SpeechLM models on SUPERB \citep{yang2021superb}, which is designed to provide a standard and comprehensive testbed for pre-trained models on various speech tasks, including Speaker Identification (SID),	 Automatic Speaker Verification (ASV), Speaker Diarization (SD), Phoneme Recognition (PR), Automatic Speech Recognition (ASR),  Out-Of-Domain Automatic Speech Recognition (OOD-ASR), Keyword Spotting (KS), Query by Example Spoken Term Detection (QbE),  Speech Translation (ST), Intent Classification (IC), Slot Filling (SF), Emotion Recognition (ER). These tasks can be grouped into five aspects of speech: content, speaker, semantics, and paralinguistics (ParaL). Table~\ref{table:superb} shows the universal speech representation evaluation results. Compared to the previous self-supervised learning methods, SpeechLM achieves good performance on several content-related and semantic-related tasks, such as PR, ASR, ST, and SF. Particularly, the proposed SpeechLM-P model obtains 36\% and 20\% relative PER/WER reductions on PR and ASR tasks. Meanwhile, we can observe performance degradation for the speaker and paralinguistics-related tasks, especially for SpeechLM-P. It indicates that with our joint speech and text pre-training method, the model learns more about extracting the content-related information while discarding the other aspects of speech signals.

\subsection{Analysis}
To better understand the effectiveness of the proposed method, we conduct several experiments to investigate its main components, such as the random swapping mechanism, the comparison of two tokenizers, the amounts of unpaired text data, and further visualization analysis\footnote{
More ablations such as the effect of the speech/text pre-training ratio could be found in Appendix \ref{Appen:Analysis}.
}.

\begin{table*}[!ht]
\begin{center}
\resizebox{\linewidth}{!}{
\renewcommand\arraystretch{1.1}
\begin{tabular}{l|lc|ccc|cc|ccc}
\hline
\multicolumn{1}{c|}{\multirow{2}{*}{\#}}     & \multirow{2}{*}{Model}        & \multirow{2}{*}{Size}  & \multicolumn{3}{c|}{Pre-training Data}       & \multicolumn{2}{c|}{WER ($\downarrow$) w/o LM} & \multicolumn{3}{c}{WER ($\downarrow$) w/ LM}   \\
                        &                               &                        & Speech           & Paired        & Text      & test-clean    & test-other    & LM        & test-clean    & test-other \\
\hline
1 & SpeechLM-P                                          & Base (0.1B)            & 960h             & 100h          & 40M       & 3.4           & 8.1          & 4-gram    & 2.7           & 6.2   \\ 
2 & SpeechLM-P w/o swapping                             & Base (0.1B)            & 960h             & 100h          & 40M       & 4.0           & 9.1           & 4-gram    & 2.9           & 6.7  \\ 
\hline
3 & SpeechLM-P w/o text pre-training                    & Base (0.1B)            & 960h             & 100h          & -         & 4.9           & 10.4          & 4-gram    & 3.0           & 7.0 \\
4 & SpeechLM-H w/o text pre-training                    & Base (0.1B)            & 960h             & -             & -         & 4.7           & 10.1          & 4-gram    & 3.0           & 6.9 \\
\hdashline
5 & SpeechLM-H                                          & Base (0.1B)            & 960h             & 100h          & 400K       & 3.8           & 8.3           & 4-gram    & 2.7           & 6.0   \\ 
6 & SpeechLM w/o paired data                            & Base (0.1B)            & 960h             & -             & 40M       & 4.5           & 9.9           & 4-gram    & 3.0           & 6.9   \\ 
\hline
\end{tabular}
}
\caption{\label{Tab:ablations} Ablation study on 100-hour LibriSpeech benchmark. The paired data are used for training tokenizers.
}
\end{center}
\end{table*}

\begin{figure*}[!ht]
    \centering
    \includegraphics[width=0.95\textwidth]{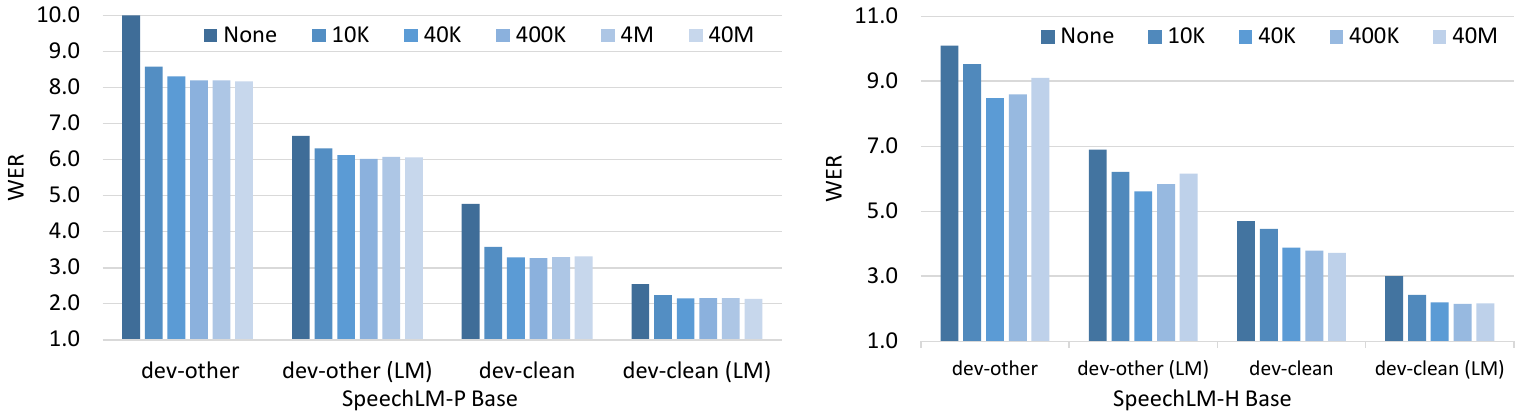}
    \caption{ASR performance fine-tuned on 100-hour LibriSpeech benchmark, models are pre-trained with different amounts of text data.
    }\label{fig:data_size_ablation}
    \vspace{0.3cm}
\end{figure*}%

\begin{figure*}[!ht]
	\centering
	\includegraphics[width=1.0\textwidth]{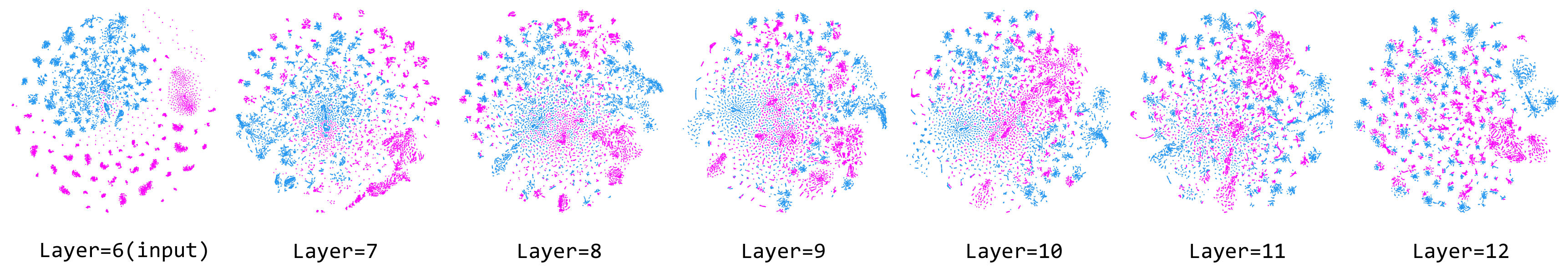}
	\caption{Layer-wise visualization of the Shared Transformer in SpeechLM-P Base. Frame-wise representations of unpaired speech (blue) and phonemes (red) are present.}\label{fig:visualization}
\end{figure*}

\paragraph{Effect of Random Swapping Mechanism}
The proposed random swapping mechanism is the key component of SpeechLM to align the speech and text modalities in the same space. Here, we explore its effectiveness by removing it. 
As shown in lines 1-2 of Table \ref{Tab:ablations}, without the random swapping mechanism, the performance declines dramatically from 8.1 WER to 9.1 WER in \texttt{test-other} set without LM, and it confirms our suspicions.

\paragraph{Comparison of Two Tokenizers}
To further compare the influence of the two tokenizers, we pre-train two models with only speech data, with results shown in lines 3-4 of Table \ref{Tab:ablations}.
Lines 3-4 show that two tokenizers perform comparably for downstream ASR tasks.
Moreover, we explore whether we can obtain improvement by not relying on paired speech-text data for training tokenizers.
We conduct an experiment (line 6) in which the speech side predicts the HuBERT hidden units and the text side is trained with masked phoneme-to-character CTC loss.
Compared to the results using pair data (line 5), the performance is degraded drastically, indicating the paired data are necessary for aligning the modalities.

\paragraph{Effect of Text Data Size} \label{ssec:analysis}
Since the text corpus contains up to 40M sentences which is much larger than the number of speech samples (960-hour Librispeech contains about 30K sentences), we conduct experiments to explore the effect of text data size for pre-training, by randomly sampling subsets from the original text corpus.
Surprisingly, Figure \ref{fig:data_size_ablation} shows that the performance does not degrade much until the text data are reduced to 40K sentences.
We speculate that the text data here are modeled at the lexical level, i.e., the transformation from phoneme/hidden units to characters, and 40K data is sufficient to build a lexicon.
It is also noted that the WER of \texttt{dev-other} set is getting worse as the amount of text data increases for the SpeechLM-H models, while such degradation is not observed for SpeechLM-P.
It is possible due to the hidden-unit tokenizer $\mathcal{T}^H_T$ trained on 100h clean unit-to-text data, since the tokenization errors can accumulate as the amount of text data increases.

\paragraph{Visualization Analysis}
Figure \ref{fig:visualization} illustrates the data distributions from different layers of the Shared Transformer in the SpeechLM-P Base model.
The dimension is reduced to 2-D by T-SNE \citep{van2008visualizing}.
Data points are randomly sampled from unpaired speech and text samples from LibriSpeech \texttt{dev-clean} set.
\texttt{Layer=6} denotes the input.
It is shown that as the layer increases, SpeechLM is able to align speech and text representations into a shared space.

\section{Conclusion}
In this work, we present SpeechLM, a text-augmented speech pre-trained model, which achieves competitive performance on various spoken language tasks, such as automatic speech recognition and speech translation. To make full use of unpaired data, we propose two alternative discrete tokenizers based on phoneme units and hidden units to tokenize speech and text into the same semantic space.
With the shared interface, SpeechLM can learn better speech representations with the help of text modality.
Quantitative and qualitative analyses demonstrate the superiority and effectiveness of the proposed method.
For future work, we would like to advance the work by deeply integrating the language model ability and extending to natural language tasks.

\section*{Limitations}
While the proposed SpeechLM achieves competitive performance on various spoken language tasks, it still has some limitations: (1) the current method needs paired data, or phoneme lexicon to build the tokenizers. The lexicon might be language-specific, which restricts the cross/multi-lingual application; (2) the effectiveness of applying SpeechLM to other speech domains (e.g., noisy, conversation-style speech) and the minimum amount of paired data required to build well-performing tokenizers need to be further investigated; (3) due to our computation limits, the performance of SpeechLM X-Large models are not explored.

\section*{Ethics Statement}
This work presents a text-augmented speech pre-trained model SpeechLM.
We evaluate our methods on standard benchmarks of the research community.
The datasets used in this study contain LibriSpeech \cite{panayotov2015librispeech}, LibriLight \cite{kahn2020librilight}, LibriSpeech LM Corpus \cite{panayotov2015librispeech}, and CoVoST \cite{wang2020covost}.
And the SUPERB benchmark is from \citet{yang2021superb}.
They are all public datasets or benchmarks that are widely used in the research community.

\bibliography{custom}

\begin{thebibliography}{44}
\expandafter\ifx\csname natexlab\endcsname\relax\def\natexlab#1{#1}\fi

\bibitem[{Ao et~al.(2022)Ao, Wang, Zhou, Wang, Ren, Wu, Liu, Ko, Li, Zhang,
  Wei, Qian, Li, and Wei}]{ao2021speecht5}
Junyi Ao, Rui Wang, Long Zhou, Chengyi Wang, Shuo Ren, Yu~Wu, Shujie Liu, Tom
  Ko, Qing Li, Yu~Zhang, Zhihua Wei, Yao Qian, Jinyu Li, and Furu Wei. 2022.
\newblock \href {https://doi.org/10.18653/v1/2022.acl-long.393} {{S}peech{T}5:
  Unified-modal encoder-decoder pre-training for spoken language processing}.
\newblock In \emph{Proceedings of the 60th Annual Meeting of the Association
  for Computational Linguistics (Volume 1: Long Papers)}, pages 5723--5738,
  Dublin, Ireland. Association for Computational Linguistics.

\bibitem[{Baevski et~al.(2021)Baevski, Hsu, CONNEAU, and
  Auli}]{NEURIPS2021_w2vu}
Alexei Baevski, Wei-Ning Hsu, Alexis CONNEAU, and Michael Auli. 2021.
\newblock \href
  {https://proceedings.neurips.cc/paper/2021/file/ea159dc9788ffac311592613b7f71fbb-Paper.pdf}
  {Unsupervised speech recognition}.
\newblock In \emph{Advances in Neural Information Processing Systems},
  volume~34, pages 27826--27839. Curran Associates, Inc.

\bibitem[{Baevski et~al.(2022)Baevski, Hsu, Xu, Babu, Gu, and
  Auli}]{baevski2022data2vec}
Alexei Baevski, Wei-Ning Hsu, Qiantong Xu, Arun Babu, Jiatao Gu, and Michael
  Auli. 2022.
\newblock Data2vec: A general framework for self-supervised learning in speech,
  vision and language.
\newblock \emph{arXiv preprint arXiv:2202.03555}.

\bibitem[{Baevski et~al.(2020{\natexlab{a}})Baevski, Schneider, and
  Auli}]{vq_wav2vec}
Alexei Baevski, Steffen Schneider, and Michael Auli. 2020{\natexlab{a}}.
\newblock vq-wav2vec: Self-supervised learning of discrete speech
  representations.
\newblock In \emph{International Conference on Learning Representations
  (ICLR)}.

\bibitem[{Baevski et~al.(2020{\natexlab{b}})Baevski, Zhou, Mohamed, and
  Auli}]{wav2vec2}
Alexei Baevski, Yuhao Zhou, Abdelrahman Mohamed, and Michael Auli.
  2020{\natexlab{b}}.
\newblock wav2vec 2.0: {A} framework for self-supervised learning of speech
  representations.
\newblock In \emph{Advances in Neural Information Processing Systems
  (NeurIPS)}.

\bibitem[{Bapna et~al.(2022)Bapna, Cherry, Zhang, Jia, Johnson, Cheng, Khanuja,
  Riesa, and Conneau}]{bapna2022mslam}
Ankur Bapna, Colin Cherry, Yu~Zhang, Ye~Jia, Melvin Johnson, Yong Cheng, Simran
  Khanuja, Jason Riesa, and Alexis Conneau. 2022.
\newblock mslam: Massively multilingual joint pre-training for speech and text.
\newblock \emph{arXiv preprint arXiv:2202.01374}.

\bibitem[{Bapna et~al.(2021)Bapna, Chung, Wu, Gulati, Jia, Clark, Johnson,
  Riesa, Conneau, and Zhang}]{bapna2021slam}
Ankur Bapna, Yu-an Chung, Nan Wu, Anmol Gulati, Ye~Jia, Jonathan~H Clark,
  Melvin Johnson, Jason Riesa, Alexis Conneau, and Yu~Zhang. 2021.
\newblock Slam: A unified encoder for speech and language modeling via
  speech-text joint pre-training.
\newblock \emph{arXiv preprint arXiv:2110.10329}.

\bibitem[{Chen et~al.(2022{\natexlab{a}})Chen, Wang, Chen, Wu, Liu, Chen, Li,
  Kanda, Yoshioka, Xiao et~al.}]{wavlm}
Sanyuan Chen, Chengyi Wang, Zhengyang Chen, Yu~Wu, Shujie Liu, Zhuo Chen, Jinyu
  Li, Naoyuki Kanda, Takuya Yoshioka, Xiong Xiao, et~al. 2022{\natexlab{a}}.
\newblock Wavlm: Large-scale self-supervised pre-training for full stack speech
  processing.
\newblock \emph{IEEE Journal of Selected Topics in Signal Processing}.

\bibitem[{Chen et~al.(2022{\natexlab{b}})Chen, Zhang, Rosenberg, Ramabhadran,
  Moreno, Bapna, and Zen}]{chen2022maestro}
Zhehuai Chen, Yu~Zhang, Andrew Rosenberg, Bhuvana Ramabhadran, Pedro Moreno,
  Ankur Bapna, and Heiga Zen. 2022{\natexlab{b}}.
\newblock Maestro: Matched speech text representations through modality
  matching.
\newblock \emph{arXiv preprint arXiv:2204.03409}.

\bibitem[{Chung et~al.(2019)Chung, Hsu, Tang, and Glass}]{apc1}
Yu-An Chung, Wei-Ning Hsu, Hao Tang, and James Glass. 2019.
\newblock {An Unsupervised Autoregressive Model for Speech Representation
  Learning}.
\newblock In \emph{Interspeech}, pages 146--150.

\bibitem[{Chung et~al.(2020)Chung, Tang, and Glass}]{vq_apc}
Yu-An Chung, Hao Tang, and James Glass. 2020.
\newblock Vector-quantized autoregressive predictive coding.
\newblock In \emph{Interspeech}, pages 3760--3764.

\bibitem[{Chung et~al.(2021)Chung, Zhang, Han, Chiu, Qin, Pang, and
  Wu}]{chung2021w2v}
Yu-An Chung, Yu~Zhang, Wei Han, Chung-Cheng Chiu, James Qin, Ruoming Pang, and
  Yonghui Wu. 2021.
\newblock W2v-bert: Combining contrastive learning and masked language modeling
  for self-supervised speech pre-training.
\newblock In \emph{2021 IEEE Automatic Speech Recognition and Understanding
  Workshop (ASRU)}, pages 244--250. IEEE.

\bibitem[{Devlin et~al.(2019)Devlin, Chang, Lee, and
  Toutanova}]{devlin2018bert}
Jacob Devlin, Ming-Wei Chang, Kenton Lee, and Kristina Toutanova. 2019.
\newblock \href {https://doi.org/10.18653/v1/N19-1423} {{BERT}: Pre-training of
  deep bidirectional transformers for language understanding}.
\newblock In \emph{Proceedings of the 2019 Conference of the North {A}merican
  Chapter of the Association for Computational Linguistics: Human Language
  Technologies, Volume 1 (Long and Short Papers)}, pages 4171--4186.

\bibitem[{Dong et~al.(2019)Dong, Yang, Wang, Wei, Liu, Wang, Gao, Zhou, and
  Hon}]{dong2019unified}
Li~Dong, Nan Yang, Wenhui Wang, Furu Wei, Xiaodong Liu, Yu~Wang, Jianfeng Gao,
  Ming Zhou, and Hsiao-Wuen Hon. 2019.
\newblock \href
  {https://proceedings.neurips.cc/paper/2019/file/c20bb2d9a50d5ac1f713f8b34d9aac5a-Paper.pdf}
  {Unified language model pre-training for natural language understanding and
  generation}.
\newblock In \emph{Proceedings of the 33rd Conference on Neural Information
  Processing Systems}, volume~32, pages 13063--13075.

\bibitem[{Graves et~al.(2006)Graves, Fern\'{a}ndez, Gomez, and
  Schmidhuber}]{Graves10.1145CTC}
Alex Graves, Santiago Fern\'{a}ndez, Faustino Gomez, and J\"{u}rgen
  Schmidhuber. 2006.
\newblock \href {https://doi.org/10.1145/1143844.1143891} {Connectionist
  temporal classification: Labelling unsegmented sequence data with recurrent
  neural networks}.
\newblock ICML '06, page 369–376, New York, NY, USA. Association for
  Computing Machinery.

\bibitem[{Hsu et~al.(2021)Hsu, Bolte, Tsai, Lakhotia, Salakhutdinov, and
  Mohamed}]{hsu2021hubert}
Wei-Ning Hsu, Benjamin Bolte, Yao-Hung~Hubert Tsai, Kushal Lakhotia, Ruslan
  Salakhutdinov, and Abdelrahman Mohamed. 2021.
\newblock \href {https://doi.org/10.1109/TASLP.2021.3122291} {Hubert:
  Self-supervised speech representation learning by masked prediction of hidden
  units}.
\newblock \emph{IEEE/ACM Transactions on Audio, Speech, and Language
  Processing}, 29:3451--3460.

\bibitem[{Joshi et~al.(2020)Joshi, Chen, Liu, Weld, Zettlemoyer, and
  Levy}]{joshi2020spanbert}
Mandar Joshi, Danqi Chen, Yinhan Liu, Daniel~S Weld, Luke Zettlemoyer, and Omer
  Levy. 2020.
\newblock Spanbert: Improving pre-training by representing and predicting
  spans.
\newblock \emph{Transactions of the Association for Computational Linguistics},
  8:64--77.

\bibitem[{Kahn et~al.(2020)Kahn, Rivi{\`e}re, Zheng, Kharitonov, Xu,
  Mazar{\'e}, Karadayi, Liptchinsky, Collobert, Fuegen
  et~al.}]{kahn2020librilight}
Jacob Kahn, Morgane Rivi{\`e}re, Weiyi Zheng, Evgeny Kharitonov, Qiantong Xu,
  Pierre-Emmanuel Mazar{\'e}, Julien Karadayi, Vitaliy Liptchinsky, Ronan
  Collobert, Christian Fuegen, et~al. 2020.
\newblock \href {https://doi.org/10.1109/ICASSP40776.2020.9052942}
  {Libri-light: A benchmark for asr with limited or no supervision}.
\newblock In \emph{Proceedings of the 2020 IEEE International Conference on
  Acoustics, Speech and Signal Processing}, pages 7669--7673. IEEE.

\bibitem[{Kim et~al.(2021)Kim, Kim, Lee, and Ha}]{kim2021st}
Minjeong Kim, Gyuwan Kim, Sang-Woo Lee, and Jung-Woo Ha. 2021.
\newblock \href {https://doi.org/10.1109/ICASSP39728.2021.9414558} {St-bert:
  Cross-modal language model pre-training for end-to-end spoken language
  understanding}.
\newblock In \emph{Proceedings of the 2021 IEEE International Conference on
  Acoustics, Speech and Signal Processing}, pages 7478--7482.

\bibitem[{Kingma and Ba(2014)}]{kingma2014adam}
Diederik~P Kingma and Jimmy Ba. 2014.
\newblock Adam: A method for stochastic optimization.
\newblock \emph{arXiv preprint arXiv:1412.6980}.

\bibitem[{Ling and Liu(2020)}]{decoar2}
Shaoshi Ling and Yuzong Liu. 2020.
\newblock {DeCoAR} 2.0: Deep contextualized acoustic representations with
  vector quantization.
\newblock \emph{arXiv preprint arXiv:2012.06659}.

\bibitem[{Liu et~al.(2020{\natexlab{a}})Liu, Chung, and Glass}]{npc}
Alexander~H Liu, Yu-An Chung, and James Glass. 2020{\natexlab{a}}.
\newblock Non-autoregressive predictive coding for learning speech
  representations from local dependencies.
\newblock \emph{arXiv preprint arXiv:2011.00406}.

\bibitem[{Liu et~al.(2020{\natexlab{b}})Liu, Li, and Lee}]{tera}
Andy~T Liu, Shang-Wen Li, and Hung-yi Lee. 2020{\natexlab{b}}.
\newblock Tera: Self-supervised learning of transformer encoder representation
  for speech.
\newblock \emph{arXiv preprint arXiv:2007.06028}.

\bibitem[{Liu et~al.(2020{\natexlab{c}})Liu, Yang, Chi, Hsu, and
  Lee}]{mockingjay}
Andy~T. Liu, Shu-wen Yang, Po-Han Chi, Po-chun Hsu, and Hung-yi Lee.
  2020{\natexlab{c}}.
\newblock Mockingjay: Unsupervised speech representation learning with deep
  bidirectional transformer encoders.
\newblock In \emph{International Conference on Acoustics, Speech and Signal
  Processing (ICASSP)}. IEEE.

\bibitem[{Mohri et~al.(2002)Mohri, Pereira, and Riley}]{mohri2002weighted}
Mehryar Mohri, Fernando Pereira, and Michael Riley. 2002.
\newblock Weighted finite-state transducers in speech recognition.
\newblock \emph{Computer Speech \& Language}, 16(1):69--88.

\bibitem[{Panayotov et~al.(2015)Panayotov, Chen, Povey, and
  Khudanpur}]{panayotov2015librispeech}
Vassil Panayotov, Guoguo Chen, Daniel Povey, and Sanjeev Khudanpur. 2015.
\newblock \href {https://doi.org/10.1109/ICASSP.2015.7178964} {Librispeech: an
  asr corpus based on public domain audio books}.
\newblock In \emph{Proceedings of the 2015 IEEE International Conference on
  Acoustics, Speech and Signal Processing}, pages 5206--5210. IEEE.

\bibitem[{Papineni et~al.(2002)Papineni, Roukos, Ward, and
  Zhu}]{papineni2002bleu}
Kishore Papineni, Salim Roukos, Todd Ward, and Wei-Jing Zhu. 2002.
\newblock Bleu: a method for automatic evaluation of machine translation.
\newblock In \emph{Proceedings of the 40th annual meeting of the Association
  for Computational Linguistics}, pages 311--318.

\bibitem[{Povey et~al.(2011)Povey, Ghoshal, Boulianne, Burget, Glembek, Goel,
  Hannemann, Motlicek, Qian, Schwarz, Silovsky, Stemmer, and Vesely}]{kaldi}
Daniel Povey, Arnab Ghoshal, Gilles Boulianne, Lukas Burget, Ondrej Glembek,
  Nagendra Goel, Mirko Hannemann, Petr Motlicek, Yanmin Qian, Petr Schwarz, Jan
  Silovsky, Georg Stemmer, and Karel Vesely. 2011.
\newblock \href {http://infoscience.epfl.ch/record/192584} {The kaldi speech
  recognition toolkit}.
\newblock IEEE Signal Processing Society.
\newblock IEEE Catalog No.: CFP11SRW-USB.

\bibitem[{Qian et~al.(2021)Qian, Bianv, Shi, Kanda, Shen, Xiao, and
  Zeng}]{qian2021speech}
Yao Qian, Ximo Bianv, Yu~Shi, Naoyuki Kanda, Leo Shen, Zhen Xiao, and Michael
  Zeng. 2021.
\newblock \href {https://doi.org/10.1109/ICASSP39728.2021.9414900}
  {Speech-language pre-training for end-to-end spoken language understanding}.
\newblock In \emph{Proceedings of the 2021 IEEE International Conference on
  Acoustics, Speech and Signal Processing}, pages 7458--7462.

\bibitem[{Ravanelli et~al.(2020)Ravanelli, Zhong, Pascual, Swietojanski,
  Monteiro, Trmal, and Bengio}]{pase+}
Mirco Ravanelli, Jianyuan Zhong, Santiago Pascual, Pawel Swietojanski, Joao
  Monteiro, Jan Trmal, and Yoshua Bengio. 2020.
\newblock Multi-task self-supervised learning for robust speech recognition.
\newblock In \emph{International Conference on Acoustics, Speech and Signal
  Processing (ICASSP)}, pages 6989--6993. IEEE.

\bibitem[{Ren et~al.(2019)Ren, Ruan, Tan, Qin, Zhao, Zhao, and
  Liu}]{ren2019fastspeech}
Yi~Ren, Yangjun Ruan, Xu~Tan, Tao Qin, Sheng Zhao, Zhou Zhao, and Tie-Yan Liu.
  2019.
\newblock Fastspeech: Fast, robust and controllable text to speech.
\newblock \emph{Advances in Neural Information Processing Systems}, 32.

\bibitem[{Rivi{\`e}re et~al.(2020)Rivi{\`e}re, Joulin, Mazar{\'e}, and
  Dupoux}]{modified_cpc}
Morgane Rivi{\`e}re, Armand Joulin, Pierre-Emmanuel Mazar{\'e}, and Emmanuel
  Dupoux. 2020.
\newblock Unsupervised pretraining transfers well across languages.
\newblock In \emph{International Conference on Acoustics, Speech and Signal
  Processing (ICASSP)}, pages 7414--7418. IEEE.

\bibitem[{Schneider et~al.(2019)Schneider, Baevski, Collobert, and
  Auli}]{schneider2019wav2vec}
Steffen Schneider, Alexei Baevski, Ronan Collobert, and Michael Auli. 2019.
\newblock wav2vec: Unsupervised pre-training for speech recognition.
\newblock \emph{arXiv preprint arXiv:1904.05862}.

\bibitem[{Shaw et~al.(2018)Shaw, Uszkoreit, and Vaswani}]{shaw-etal-2018-self}
Peter Shaw, Jakob Uszkoreit, and Ashish Vaswani. 2018.
\newblock \href {https://doi.org/10.18653/v1/N18-2074} {Self-attention with
  relative position representations}.
\newblock In \emph{Proceedings of the 2018 Conference of the North American
  Chapter of the Association for Computational Linguistics: Human Language
  Technologies, Volume 2 (Short Papers)}, pages 464--468.

\bibitem[{Tang et~al.(2022)Tang, Gong, Dong, Wang, Hsu, Gu, Baevski, Li,
  Mohamed, Auli, and Pino}]{tang2022unified}
Yun Tang, Hongyu Gong, Ning Dong, Changhan Wang, Wei-Ning Hsu, Jiatao Gu,
  Alexei Baevski, Xian Li, Abdelrahman Mohamed, Michael Auli, and Juan Pino.
  2022.
\newblock \href {https://doi.org/10.18653/v1/2022.acl-long.105} {Unified
  speech-text pre-training for speech translation and recognition}.
\newblock In \emph{Proceedings of the 60th Annual Meeting of the Association
  for Computational Linguistics (Volume 1: Long Papers)}, pages 1488--1499,
  Dublin, Ireland. Association for Computational Linguistics.

\bibitem[{Tremblay and Dick(2016)}]{tremblay2016broca}
Pascale Tremblay and Anthony~Steven Dick. 2016.
\newblock Broca and wernicke are dead, or moving past the classic model of
  language neurobiology.
\newblock \emph{Brain and language}, 162:60--71.

\bibitem[{Van~der Maaten and Hinton(2008)}]{van2008visualizing}
Laurens Van~der Maaten and Geoffrey Hinton. 2008.
\newblock Visualizing data using t-sne.
\newblock \emph{Journal of machine learning research}, 9(11).

\bibitem[{Vaswani et~al.(2017)Vaswani, Shazeer, Parmar, Uszkoreit, Jones,
  Gomez, Kaiser, and Polosukhin}]{vaswani2017attention}
Ashish Vaswani, Noam Shazeer, Niki Parmar, Jakob Uszkoreit, Llion Jones,
  Aidan~N Gomez, {\L}ukasz Kaiser, and Illia Polosukhin. 2017.
\newblock \href {https://dl.acm.org/doi/pdf/10.5555/3295222.3295349} {Attention
  is all you need}.
\newblock In \emph{Proceedings of the 31st Conference on Neural Information
  Processing Systems}, volume~30, pages 6000--6010.

\bibitem[{Wang et~al.(2020)Wang, Wu, and Pino}]{wang2020covost}
Changhan Wang, Anne Wu, and Juan Pino. 2020.
\newblock Covost 2 and massively multilingual speech-to-text translation.
\newblock \emph{arXiv preprint arXiv:2007.10310}.

\bibitem[{Wang et~al.(2021)Wang, Wu, Pino, Baevski, Auli, and
  Conneau}]{Changhan2021Large}
Changhan Wang, Anne Wu, Juan~Miguel Pino, Alexei Baevski, Michael Auli, and
  Alexis Conneau. 2021.
\newblock \href {https://arxiv.org/abs/2104.06678} {Large-scale self- and
  semi-supervised learning for speech translation}.
\newblock \emph{CoRR}, abs/2104.06678.

\bibitem[{Wang et~al.(2022{\natexlab{a}})Wang, Wang, Wu, Chen, Li, Liu, and
  Wei}]{wang2022supervision}
Chengyi Wang, Yiming Wang, Yu~Wu, Sanyuan Chen, Jinyu Li, Shujie Liu, and Furu
  Wei. 2022{\natexlab{a}}.
\newblock Supervision-guided codebooks for masked prediction in speech
  pre-training.
\newblock \emph{arXiv preprint arXiv:2206.10125}.

\bibitem[{Wang et~al.(2022{\natexlab{b}})Wang, Wu, Chen, Liu, Li, Qian, and
  Yang}]{wang2021self}
Chengyi Wang, Yu~Wu, Sanyuan Chen, Shujie Liu, Jinyu Li, Yao Qian, and Zhenglu
  Yang. 2022{\natexlab{b}}.
\newblock \href {https://doi.org/10.1109/ICASSP43922.2022.9747022} {Improving
  self-supervised learning for speech recognition with intermediate layer
  supervision}.
\newblock In \emph{{IEEE} International Conference on Acoustics, Speech and
  Signal Processing, {ICASSP} 2022, Virtual and Singapore, 23-27 May 2022},
  pages 7092--7096. {IEEE}.

\bibitem[{Yang et~al.(2021)Yang, Chi, Chuang, Lai, Lakhotia, Lin, Liu, Shi,
  Chang, Lin et~al.}]{yang2021superb}
Shu-wen Yang, Po-Han Chi, Yung-Sung Chuang, Cheng-I~Jeff Lai, Kushal Lakhotia,
  Yist~Y Lin, Andy~T Liu, Jiatong Shi, Xuankai Chang, Guan-Ting Lin, et~al.
  2021.
\newblock Superb: Speech processing universal performance benchmark.
\newblock \emph{arXiv preprint arXiv:2105.01051}.

\bibitem[{Zhang et~al.(2022)Zhang, Zhou, Ao, Liu, Dai, Li, and
  Wei}]{zhang2022speechut}
Ziqiang Zhang, Long Zhou, Junyi Ao, Shujie Liu, Lirong Dai, Jinyu Li, and Furu
  Wei. 2022.
\newblock Speechut: Bridging speech and text with hidden-unit for
  encoder-decoder based speech-text pre-training.
\newblock \emph{arXiv preprint arXiv:2210.03730}.

\end{thebibliography}
\bibliographystyle{acl_natbib}

\appendix

\newpage
\section{Appendix}
\subsection{Tokenizer Details} \label{Appen:tok_detail}
\paragraph{Phone-unit tokenizer for speech}
In the Base setting, we train a hybrid GMM-HMM ASR model (\texttt{tri4b}) on 100 hours of labeled LibriSpeech data following Kaldi recipe~\citep{kaldi}.
To boost performance, we then use the \texttt{tri4b} model to decode the remaining 860 hours of speech and train the \texttt{tri6b} model on all the pseudo-labeled data, which is finally used for the phone-unit tokenizer. 
In the Large setting, we train a neural network instead of GMM with 960- hour labeled LibriSpeech data, which can boost the performance and alignment accuracy.
Once the hybrid model is trained, unlabeled speech data is decoded and transduced to the best phoneme-level alignment paths.
The frame shift is 10ms for the Base model setup, and 30ms for the Large model setup, respectively.
We then re-sample the phonemes to a frame rate of 20ms by linear interpolation.
\paragraph{Phone-unit tokenizer for text}
We use the 200K word-to-phone lexicon provided by LibriSpeech to convert words to phonemes, the OOV words are replaced by \texttt{<unk>} symbol.
Following \citet{NEURIPS2021_w2vu}, we randomly insert \texttt{<SIL>} phoneme between words with a probability of 25\%.
Then we upsample the phoneme sequence by repeating the phonemes.
The length of phonemes follows Gaussian distribution estimated from the \texttt{train} set of LibriSpeech, specifically, the mean is 5 and the variance is 25 except for the \texttt{<SIL>} phoneme which has a mean of 14.

\paragraph{Hidden-unit tokenizer for speech}
We use the released HuBERT \citep{hsu2021hubert} model following a K-Means model as the tokenizer for speech. The K-Means model has 500 classes with a frame rate of 50.

\paragraph{Hidden-unit tokenizer for text}
To build a text-to-hidden-unit tokenizer, we modify FastSpeech \citep{ren2019fastspeech} by replacing the prediction head from predicting the spectrum to predicting the probability of hidden units.
Specifically, the tokenizer has 4 layers of encoders and 4 layers of decoders, with a model dimension of 256.
The input to the model is a phoneme sequence converted from raw text.
Upsampling is performed by a duration model between the encoder and the decoder, which predicts the length of each phoneme and repeats the phonemes before feeding them into the decoder.
We train the model on LibriSpeech \texttt{train-clean-100} subset for 10K steps, with a learning rate of 5e-4 and a batch size of 10K phonemes.
The final model achieves 41.3 and 34.6 BLEU scores on \texttt{dev-clean} and \texttt{dev-other}.


\begin{table*}[!ht]
\begin{center}
\small
\begin{tabular}{c|cc|ccc}
\hline
\multirow{2}{*}{Ratio ($\lambda$)} & \multicolumn{2}{c|}{WER ($\downarrow$) w/o LM} & \multicolumn{3}{c}{WER ($\downarrow$) With LM} \\
                                   & dev-clean                & dev-other               & LM           & dev-clean      & dev-other      \\ \hline
0.01                               & 3.67                     & 8.35                    & 4-gram       & 2.24           & 6.19           \\
0.1                                & 3.32                     & \textbf{8.17}           & 4-gram       & \textbf{2.14}  & \textbf{6.06}   \\
1.0                                & \textbf{3.18}            & 8.35                    & 4-gram       & 2.16           & 6.23           \\
10.0                               & 3.52                     & 9.55                    & 4-gram       & 2.31           & 7.20           \\ \hline
\end{tabular}
\caption{\label{Tab: loss_ratio} ASR performance on 100-hour LibriSpeech benchmark. Different ratio of text pre-training loss in SpeechLM-P model.}
\end{center}
\end{table*}

\begin{table*}[ht!]
\begin{center}
\resizebox{\linewidth}{!}{
\renewcommand\arraystretch{1.0}
\begin{tabular}{lc|ccc|cc|ccc}
\hline
\multirow{2}{*}{Model}                  & \multirow{2}{*}{Size}  & \multicolumn{3}{c|}{Pre-training Data}        & \multicolumn{2}{c|}{WER ($\downarrow$) w/o LM} & \multicolumn{3}{c}{WER ($\downarrow$) w/ LM}   \\
                                        &                        & Speech           & Paired         & Text      & test-clean          & test-other         & LM        & test-clean& test-other \\
\hline
SLAM \citep{bapna2021slam}             & X-Large (0.6B)        & 60kh        & 960h        & mC4-En                & 1.6                & 3.1               & -            & -              & -   \\
Maestro \citep{chen2022maestro}        & X-Large (0.6B)        & 60kh        & $\sim$5kh        & 40M+TEDLIUM           & 1.5                & 2.8               & Conformer       & 1.5            & 2.7   \\
SpeechLM-P (ours)                      & Large (0.3B)          & 60kh        & 960h        & 40M                   & 1.9               & 3.6               & Transformer      & 1.8            & 3.2   \\
\hline
\end{tabular}
}
\caption{ASR performance (WER) of on 960h LibriSpeech benchmark, comparing SpeechLM with SLAM and Maestro.
}
\label{Tab: wer_xlarge} 
\end{center}
\end{table*}

\subsection{Experimental details} \label{exp_detail}
\paragraph{Pre-training configuration}
The Base model has 12 Transformer layers with the attention dimension of 768 and attention heads of 12, the Large model has 24 Transformer layers with the attention dimension of 1024 and attention heads of 16.
The convolutional layers have 512 channels and kernel sizes of [10,3,3,3,3,2,2], resulting in a downsampling rate of 320.
The CTC layer is a single 1-D convolutional layer with a kernel size of 2, whose channel matches the Transformer dimension.
It is then followed by a linear projection to the text characters.
All models are pre-trained on 32 GPUs for 400K steps including 32K warming-up steps.
We use Adam \citep{kingma2014adam} with $\beta_1$=0.9, $\beta_2$=0.98 for optimization.
The maximum learning rate is set to $5e-4$ and decays linearly to zero after the warming-up steps.

\paragraph{Fine-tuning configuration}
For Base models fine-tuned on 100-hour LibriSpeech, the total steps are 30K with a batch size of 800 seconds.
For Large models fine-tuned on the full LibriSpeech, the total steps are 200K with a batch size of 1800 seconds.
All LibriSpeech models are tuned with a maximum learning rate of 1e-5 and a tri-stage learning rate schedule with the warming-up, holding, and decay periods of $[0.1, 0.4, 0.5]$.
And for CoVoST-2, both the Base and the Large models are fine-tuned for 50K steps with a batch size of 1600 seconds.
The learning rate warms up to 1e-4 in 5K steps and then decays linearly to zero. 
After fine-tuning, we select the model with the best accuracy on the valid set in the Base setting and average the top 5 models with the best accuracy on the valid set in the Large setting.
The decoding beam size is 5 without external language model fusion.

\subsection{Analysis}   \label{Appen:Analysis}


\paragraph{Effect of Speech/Text Pre-Training Ratio} \label{Appen:pretraining_ratio}
Table \ref{Tab: loss_ratio} shows the fine-tuning performance of the different pre-trained models with respect to the pre-training loss ratio $\lambda$.
It is noticed that a lower weight (0.1) of the text pre-training task achieves the best performance in the dev set. Hence, we use $\lambda=0.1$ for all other experiments.

\subsection{ASR results on 960h LibriSpeech benchmark}   \label{Appen:XLarge-asr}
Table \ref{Tab: wer_xlarge} lists the ASR performance of SpeechLM on the full 960h LibriSpeech benchmark comparing with SLAM \citep{bapna2021slam} and Maestro \citep{chen2022maestro}. Note that they use 2$\times$ model size, a larger amount of paired data, or a different inference framework (e.g., RNN-T in MAESTRO), making the results not fairly comparable with SpeechLM.

\end{document}